  \documentclass{gretsi}
 \usepackage[ansinew]{inputenc}  
\usepackage{amsmath,graphicx}
\usepackage{amssymb,bm}
\usepackage{float}
\usepackage{colortbl}
\usepackage{subfigure}
\usepackage[english]{babel}
\usepackage{amsthm}
\usepackage[dvipsnames]{xcolor}
\definecolor{mygreen}{RGB}{0, 199, 0}
\definecolor{myorange}{RGB}{250, 100, 0}
\definecolor{myred}{RGB}{200, 0, 0}
\definecolor{myblue}{RGB}{30, 144, 255}
\definecolor{mylightskyblue}{RGB}{135, 206, 250}
\definecolor{myskyblue}{RGB}{0, 191, 255}
\definecolor{mypowderblue}{RGB}{176, 196, 222}
\newtheorem{theorem}{Théorème}[section]
\newtheorem{proposition}[theorem]{Proposition}
\newtheorem{remark}[theorem]{Remarque}
 
\titre{FEMDA : une m\'ethode de classification robuste et flexible}

\auteur{\coord{Pierre}{HOUDOUIN}{1},
    \coord{Matthieu}{JONCKHEERE}{2}, \coord{Fr\'ed\'eric}{PASCAL}{1}}

\adresse{\affil{1}{Universit\'e Paris-Saclay, CNRS, CentraleSup\'elec,
Laboratoire des signaux et syst\`emes \\
91190, Gif-sur-Yvette, France \\}
         \affil{2}{LAAS-CNRS, Universit\'e de Toulouse,
31077 Toulouse, France}}

\email{pierre.houdouin@centralesupelec.fr,
frederic.pascal@centralesupelec.fr\\
mjonckheer@laas.fr}

\resumefrancais{L'analyse discriminante quadratique (QDA) est une m\'ethode de classification qui suppose que les donn\'ees sont issues de distributions gaussiennes. Cette hypoth\`ese la rend peu robuste aux donn\'ees bruit\'ees. Le but de cet article est d'\'etudier la robustesse aux changements d'\'echelle dans les donn\'ees d'une nouvelle m\'ethode d'analyse discriminante o\`u chaque point est mod\'elis\'e par sa propre distribution elliptique sym\'etrique avec son propre facteur d'\'echelle. Une telle mod\'elisation fournit une grande flexibilit\'e pour traiter des donn\'ees h\'et\'erog\`enes et non identiquement distribu\'ees. Cette nouvelle m\'ethode s'av\`ere plus robuste aux changements d'\'echelle dans les donn\'ees que les autres m\'ethodes de l'\'etat de l'art.
}

\resumeanglais{Linear and Quadratic Discriminant Analysis (LDA and QDA) are well-known classical methods but can heavily suffer from non-Gaussian distributions and/or contaminated datasets, mainly because of the underlying Gaussian assumption that is not robust. This paper studies the robustness to scale changes in the data of a new discriminant analysis technique where each data point is drawn by its own arbitrary Elliptically Symmetrical (ES) distribution and its own arbitrary scale parameter. Such a model allows for possibly very heterogeneous, independent but non-identically distributed samples. The new decision rule derived is simple, fast and robust to scale changes in the data compared to others state-of-the-art methods.}

\begin{document}
\maketitle

\section{Introduction}

L'analyse discriminante est un outil tr\`es utilis\'e pour les t\^aches de classification. La m\'ethode historique \cite{1} pr\'esup\-pose que les donn\'ees sont issues de distributions gaussiennes et la r\`egle de d\'ecision consiste \`a choisir le cluster qui maximise la vraisemblance de la donn\'ee. Au d\'ebut des ann\'ees 80, \cite{2} et \cite{3} ont \'etudi\'e l'impact de la contamination et du \textit{mislabelling} sur les performances et concluent \`a une grande sensibilit\'e. Pour traiter ce probl\`eme, \cite{4} sugg\`ere l'utilisation de M-estimateurs qui sont robustes au bruit. Plus r\'ecemment, \cite{5} a propos\'e de mod\'eliser les donn\'ees par une distribution de student multivari\'ee, plus flexible. En 2015, \cite{6} g\'en\'eralise m\^eme aux distributions elliptiques sym\'etriques (ES). Cette nouvelle m\'ethode, appel\'ee \textit{Generalized} QDA (GQDA) repose sur l'estimation d'un seuil, dont la valeur varie avec la forme de la distribution. Enfin, \cite{7} a compl\'et\'e GQDA avec l'utilisation d'estimateurs robustes, pour obtenir RGQDA.

Toutes ces m\'ethodes supposent que les points d'un m\^eme cluster sont issus de la m\^eme distribution, hypoth\`ese qui n'est pas toujours valide. \cite{8}, inspir\'e par \cite{9}, propose une m\'ethode alternative qui ne suppose aucun a priori sur les distributions, et permet \`a chaque point d'\^etre issu de sa propre distribution elliptique sym\'etrique. Les points d'un m\^eme cluster ne sont pas forc\'ement identiquement distribu\'es, seulement tir\'es ind\'ependamment. La contrepartie d'une telle flexibilit\'e r\'eside dans les caract\'eristiques des clusters : au sein d'un m\^eme cluster, les points partagent seulement la m\^eme moyenne et la m\^eme matrice de dispersion. 
Nous allons \'etudier dans ce papier la robustesse aux changements d'\'echelle dans les donn\'ees de cette nouvelle m\'ethode.

Le mod\`ele est pr\'esent\'e dans la section 2, la section 3 contient des exp\'eriences sur donn\'ees simul\'ees, la section 4 les exp\'eriences sur donn\'ees r\'eelles et les conclusions et remarques sont effectu\'ees dans la section 5.

\section{FEMDA : Flexible EM-inspired discriminant Analysis}

\indent\textbf{Mod\`ele statistique:} On suppose que chaque donn\'ee $\mathbf{x}_i \in \mathbb{R}^m$, $i \in [1,n]$ est tir\'ee d'une distribution ES ind\'ependante du cluster. La moyenne et la matrice de dispersion d\'ependent du cluster auquel le point appartient tandis que le facteur d'\'echelle $\tau_{i,k}$ peut d\'ependre de l'observation \'egalement. La donn\'ee $\mathbf{x}_i$ du cluster $\mathcal{C}_k$, $k \in [1,K]$ est tir\'ee selon la densit\'e de probabilit\'e suivante :

$$
f(\mathbf{x}_i) =  A_{i} \left| \mathbf{\Sigma}_k \right|^{-\frac{1}{2}} \tau_{i,k}^{-\frac{m}{2}} g_{i} \left( \frac{(\mathbf{x}_i-\boldsymbol{\mu}_k)^T \mathbf{\Sigma}_k^{-1} (\mathbf{x}_i-\boldsymbol{\mu}_k)}{\tau_{i,k}} \right)
$$

\textbf{Expression de la log-vraisemblance et des estimateurs du maximum de vraisemblance:} Soient $\mathbf{x}_1,...,\mathbf{x}_{n_k}$ des donn\'ees ind\'ependantes du cluster $\mathcal{C}_k$, la log-vraisemblance de l'\'echantillon peut s'\'ecrire:
\begin{equation}\label{log-likelihood}
l(\mathbf{x}_1,...,\mathbf{x}_{n_k})
    = \sum_{i=1}^{n_k} \log \left( A_i \left| \mathbf{\Sigma}_k \right|^{-\frac{1}{2}} t_{i,k}^{-\frac{m}{2}}\, s_{i,k}^{\frac{m}{2}}\, g_i(s_{i,k}) \right) 
\end{equation}
o\`u $t_{i,k} = (\mathbf{x}_i-\boldsymbol{\mu}_k)^T \mathbf{\Sigma}_k^{-1}(\mathbf{x}_i-\boldsymbol{\mu}_k)$ and $s_{i,k}=t_{i,k}/\tau_{i,k}$. 

Maximiser le terme de l'\'equation \eqref{log-likelihood} par rapport \`a $\tau_{i,k}$, avec $\boldsymbol{\mu}_k$ et $\mathbf{\Sigma}_k$ fix\'es m\`ene \`a 
$$
\hat{\tau}_{i,k} = \frac{(\mathbf{x}_i-\boldsymbol{\mu}_k)^T \mathbf{\Sigma}_k^{-1} (\mathbf{x}_i-\boldsymbol{\mu}_k)}{\arg \max_{t \in \mathbb{R}^+} \{t^{\frac{m}{2}} g_i(t) \}}.
$$ 

Les hypoth\`eses sur $g_i$ assurent la stricte positivit\'e du d\'enominateur. Apr\`es avoir remplac\'e dans l'\'equation \eqref{log-likelihood} $\tau_{i,k}$ par $\hat{\tau}_{i,k}$, on obtient:

\begin{equation*}
l(\mathbf x_i) = \tilde{A_i} - \frac{1}{2} \log \left(\left| \mathbf{\Sigma}_k \right| \left( (\mathbf{x}_i-\boldsymbol{\mu}_k)^T \mathbf{\Sigma}_k^{-1}(\mathbf{x}_i-\boldsymbol{\mu}_k) \right)^{m}\right)   
\end{equation*}
o\`u $\tilde{A_i} = \log(A_i) + \log(\max_{t \in \mathbb{R}^+} \{t^{\frac{m}{2}} g_i(t) \}).$

A cette \'etape, on comprend que la flexibilit\'e dans le choix de l'\'echelle de la matrice de covariance nous permet de r\'eduire l'impact de la fonction g\'en\'eratrice $g_i$ dans la vraisemblance \`a une constante multiplicative ind\'ependante de $k$. Enfin, l'utilisation de l'estimateur du maximum de vraisemblance permet d'obtenir les estimateurs robustes suivants pour la moyenne et la matrice de dispersion :

\begin{equation}
\left\{\begin{array}{ccl}
\hat{\boldsymbol{\mu}}_k & = & \displaystyle \cfrac{\sum_{i=1}^{n_k} w_{i,k} \mathbf{x}_i}{\sum_{i=1}^{n_k} w_{i,k}},  \\
\hat{\mathbf{\Sigma}}_k & = & \displaystyle\frac{m}{n_k} \sum_{i=1}^{n_k} w_{i,k} (\mathbf{x}_i-\hat{\boldsymbol{\mu}}_k)(\mathbf{x}_i-\hat{\boldsymbol{\mu}}_k)^T
\end{array}
\right.
\end{equation}
o\`u $w_{i,k} = 1/t_{i,k}$.

Il est int\'eressant de noter que $\hat{\boldsymbol{\mu}}_k$ est insensible \`a l'\'echelle de $\hat{\mathbf{\Sigma}}_k$. Par cons\'equent, si $\hat{\mathbf{\Sigma}}_k$ est une solution \`a l'\'equation de point fixe, $\lambda \hat{\mathbf{\Sigma}}_k$ l'est \'egalement. Les estimateurs obtenus sont similaires aux  M-estimateurs robustes, sauf que les poids $w_{i,k}$ sont proportionnels au carr\'e de la distance de Mahalanobis. La convergence de ces deux \'equations de point-fixe coupl\'ees a \'et\'e analys\'ee par \cite{9}.

\textbf{R\`egle de classification:} Gr\^ace \`a ces deux estimateurs, on utilise les donn\'ees d'entraînement pour estimer les param\`etres inconnus. On suppose le nombre de clusters connu.
Il est maintenant possible de d\'eduire la r\`egle de classification. On a la proposition suivante :

\begin{proposition}
La r\`egle de d\'ecision pour Flexible EM-Inspired Discriminant Analysis (FEMDA) est :
\begin{equation}\label{femda_rule}
\mathbf x_i \in \mathcal{C}_k \iff \left(\forall j \neq k, \Delta_{jk}^2(\mathbf x_i) \geq \frac{1}{m} \lambda_{jk} \right)
\end{equation}

avec $\Delta_{jk}^2(\mathbf x_i) = \log \left(\cfrac{(\mathbf{x}_i- \boldsymbol{\mu}_j)^T \mathbf{\Sigma}_j^{-1}(\mathbf{x}_i- \boldsymbol{\mu}_j)}{(\mathbf{x}_i-\boldsymbol{\mu}_k)^T \mathbf{\Sigma}_k^{-1}(\mathbf{x}_i-\boldsymbol{\mu}_k)}\right)$

et $\lambda_{jk} = \log \left(\cfrac{\left| \mathbf{\Sigma}_k \right|}{\left| \mathbf{\Sigma}_j \right|} \right)$. 
\end{proposition}

\vspace{0.2cm}

\textbf{Preuve:}
La preuve repose sur le fait que la log-vraisemblance ne d\'epend de $k$ qu'\`a travers le terme $$\frac{1}{m} \log\left(\left|\mathbf{\Sigma}_k \right|\right) + \log\left((\mathbf{x}_i-\boldsymbol{\mu}_k)^T \mathbf{\Sigma}_k^{-1}(\mathbf{x}_i-\boldsymbol{\mu}_k)\right)$$

\begin{remark}
Cette r\`egle de d\'ecision est similaire \`a la version robuste de QDA. La diff\'erence est que nous comparons le logarithme des distances de Mahalanobis au carr\'e plut\^ot que directement les distances de Mahalanobis au carr\'e. Cela rend notre r\`egle de d\'ecision \'egalement insensible \`a l'\'echelle de $\mathbf{\Sigma}$.
\end{remark}

\section{Exp\'eriences sur donn\'ees simul\'ees}

FEMDA, la m\'ethode propos\'ee, est compar\'ee avec les m\'e\-thodes suivantes : QDA classique mod\'elisant les donn\'ees par des distributions gaussiennes, QDA mod\'elisant les donn\'ees par des distributions de student ($t$-QDA, \cite{5}), GQDA et GQDA \cite{6}, \cite{7}.

\textbf{Param\`etres de simulation:} Les moyennes des clusters sont tir\'ees al\'eatoirement sur la $m$-sph\`ere unit\'e. Les matrices de covariance sont g\'en\'er\'ees avec des valeurs propres et une matrice orthogonale al\'eatoires. On choisit $m=10$, $K=5$, $N_{train} = 5000$, $N_{test} = 20000$ et $\tau \sim \mathcal{U}(1, m)$.

\textbf{Sc\'enarios consid\'er\'es:} On g\'en\`ere les points gr\^ace \`a deux familles de distributions ES diff\'erentes.

\begin{center}
\begin{tabular}{|l|l|}
  \hline
  Famille de distributions & Repr\'esentation stochastique \\
  \hline
  
  gaussienne g\'en\'eralis\'ee & $\boldsymbol{\mu} + \Gamma(\frac{m}{2 \beta}, 2)^{\frac{1}{2 \beta}} \mathbf{\Sigma}^{\frac{1}{2}} \mathcal{U} \left( \mathcal{S}(0,1) \right)$ \\
  $t$-distribution & $\boldsymbol{\mu} + \mathcal{N}(0, \mathbf{\Sigma}) \sqrt{\frac{1}{\Gamma(\frac{\nu}{2}, \frac{2}{\nu})}}$ \\
  \hline
\end{tabular}
\end{center}

$\mathcal{U} \left( \mathcal{S}(0,1) \right)$ repr\'esente une distribution uniforme sur la $m$-sph\`ere unit\'e. Le param\`etre de forme $\beta$ (resp. $\nu$) est tir\'e de mani\`ere uniforme dans $[0.25, 10]$ (resp. $[1, 10]$) pour les gaussiennes g\'en\'eralis\'ees (resp. pour les $t$-distributions).

Les sc\'enarios de g\'en\'eration de donn\'ees sont d\'efinis comme suit : $0.6GG - 0.4T$ correspond \`a 60\% des points de chaque cluster g\'en\'er\'es avec une gaussienne g\'en\'eralis\'ee et 40\% avec une $t$-distribution.

On utilise le code couleur suivant pour la g\'en\'eration des param\`etres: \color{mygreen}$0.6GG - 0.4T$ \color{black} signifie que les m\^emes $\beta$ et $\nu$ sont utilis\'es pour les points d'un m\^eme cluster et \color{myred} $0.6GG - 0.4T$ \color{black} signigie qu'on utilise un param\`etre diff\'erent pour chaque point de chaque cluster.

\textbf{R\'esultats}

Pour chaque sc\'enario dans la premi\`ere colonne, le tableau \ref{tableau:1} pr\'esente les diff\'erences de taux de bonne classification entre la meilleure m\'ethode et les autres :
\vspace*{-.3cm}\begin{center}
\begin{table}[!h]
\begin{tabular}{|l|l|l|l|l|}
  \hline
  Sc\'enario & QDA & $t$-QDA & GQDA & FEMDA \\

  \hline

  GG - T &  \cellcolor{black} & \cellcolor{black} & \cellcolor{black} & \cellcolor{black} \\
      
  \hline
    
    \color{mygreen} $1-0$ \color{black} & $-0.51$ & \cellcolor{myblue} $\textbf{76.27}$ & \cellcolor{mypowderblue} $-0.47$ & \cellcolor{myskyblue} $-0.02$ \\
      
  \hline
  
    \color{mygreen} $0-1$ \color{black} & \cellcolor{mypowderblue} $-0.64$ & \cellcolor{myblue} $\textbf{76.74}$ & $-0.69$ & \cellcolor{myskyblue} $-0.16$ \\
      
  \hline
    
    \color{myred} $1-0$ \color{black} & $-0.59$ & \cellcolor{myblue} $\textbf{76.39}$ & \cellcolor{mypowderblue} $-0.58$ & \cellcolor{myskyblue} $-0.10$ \\
      
  \hline
  
    \color{myred} $0-1$ \color{black} & \cellcolor{mypowderblue} $-1.24$ & \cellcolor{myblue} $\textbf{77.08}$ & $-1.27$ & \cellcolor{myskyblue} $-0.21$ \\
      
  \hline

    \color{mygreen} $\frac{1}{2}-\frac{1}{2}$ \color{black} & $-1.17$ & \cellcolor{myblue} $\textbf{80.85}$ & \cellcolor{mypowderblue} $-1.13$ & \cellcolor{myskyblue} $-0.39$ \\
    
  \hline
    
    \color{myred} $\frac{1}{2}-\frac{1}{2}$ \color{black} & $-1.31$ & \cellcolor{myskyblue} $-0.02$ & \cellcolor{mypowderblue} $-0.87$ & \cellcolor{myblue} $\textbf{80.59}$ \\
      
  \hline

\end{tabular}
\caption{Pr\'ecision de la classification}
\label{tableau:1}
\end{table}
\end{center}
\vspace{-0.7cm}
Dans le tableau \ref{tableau:1}, on remarque que GQDA et QDA obtiennent des performances plus faibles que FEMDA et $t$-QDA. $t$-QDA est la meilleure m\'ethode dans la plupart des sc\'enarios et surpasse l\'eg\`erement FEMDA, au prix de l'estimation de plus de param\`etres et donc d'une m\'ethode plus lente. \cite{8} a \'etudi\'e plus en d\'etails les vitesses de convergence de chaque estimateur et r\`egle de d\'ecision.
Dans le tableau \ref{tableau:2}, on bruite les donn\'ees avec un changement d'\'echelle. Une fraction des donn\'ees subit le changement suivant : $x \longleftarrow \mu + \lambda (x - \mu)$. On observe alors que FEMDA est la m\'ethode la plus robuste au bruit, $t$-QDA est surpass\'ee dans presque tous les sc\'enarios lorsque la contamination atteint 25\% avec $\lambda=4$, et dans tous avec $\lambda=8$. L'\'ecart-type entre plusieurs simulations est faible, de l'ordre de $0.05\%$.

\vspace*{-.5cm}\begin{center}
\begin{table}[!h]
\begin{tabular}{|l|l|l|l|l|}
  \hline
  Sc\'enario & $t$-QDA & FEMDA & $t$-QDA & FEMDA \\
  
   \hline
  Bruit & \multicolumn{2}{l|}{10\%} & \multicolumn{2}{l|}{25\%} \\
  
   \hline

  GG - T &  \cellcolor{black} & \cellcolor{black} & \cellcolor{black} & \cellcolor{black} \\
      
  \hline
    
    \color{mygreen} $1-0$ \color{black} - $\lambda=4$ & \cellcolor{myblue} $\textbf{73.23}$ & \cellcolor{mypowderblue} $-0.19$ & \cellcolor{mypowderblue} $-0.37$ & \cellcolor{myblue} $\textbf{66.44}$ \\
      
  \hline
  
    \color{mygreen} $0-1$ \color{black} - $\lambda=4$ & \cellcolor{myblue} $\textbf{74.65}$ & \cellcolor{mypowderblue} $-0.36$ & \cellcolor{mypowderblue} $-0.15$ & \cellcolor{myblue} $\textbf{67.19}$ \\
      
  \hline
    
    \color{myred} $1-0$ \color{black} - $\lambda=4$ & \cellcolor{mypowderblue} $-0.08$ & \cellcolor{myblue} $\textbf{72.98}$ & \cellcolor{mypowderblue} $-0.22$ & \cellcolor{myblue} $\textbf{66.48}$ \\
      
  \hline
  
    \color{myred} $0-1$ \color{black} - $\lambda=4$ & \cellcolor{myblue} $\textbf{73.93}$ & \cellcolor{mypowderblue} $-0.24$ & \cellcolor{mypowderblue} $-0.04$ & \cellcolor{myblue} $\textbf{66.70}$ \\
      
  \hline

    \color{mygreen} $\frac{1}{2}-\frac{1}{2}$ \color{black} - $\lambda=4$ & \cellcolor{myblue} $\textbf{77.14}$ & \cellcolor{mypowderblue} $-0.61$ & \cellcolor{myblue} $\textbf{70.61}$ & \cellcolor{mypowderblue} $-0.21$ \\
    
  \hline
    
    \color{myred} $\frac{1}{2}-\frac{1}{2}$ \color{black} - $\lambda=4$ & \cellcolor{myblue} $\textbf{76.28}$ & \cellcolor{mypowderblue} $-0.41$ & \cellcolor{mypowderblue} $-0.13$ & \cellcolor{myblue} $\textbf{69.74}$ \\
      
  \hline
  
      \color{mygreen} $1-0$ \color{black} - $\lambda=8$ & \cellcolor{mypowderblue} $-0.11$ & \cellcolor{myblue} $\textbf{72.87}$ & \cellcolor{mypowderblue} $-0.67$ & \cellcolor{myblue} $\textbf{64.79}$ \\
      
  \hline
  
    \color{mygreen} $0-1$ \color{black} - $\lambda=8$ & \cellcolor{myblue} $\textbf{74.11}$ & \cellcolor{mypowderblue} $-0.29$ & \cellcolor{mypowderblue} $-0.45$ & \cellcolor{myblue} $\textbf{65.98}$ \\
      
  \hline
    
    \color{myred} $1-0$ \color{black} - $\lambda=8$ & \cellcolor{mypowderblue} $-0.31$ & \cellcolor{myblue} $\textbf{71.93}$ & \cellcolor{mypowderblue} $-0.33$ & \cellcolor{myblue} $\textbf{65.49}$ \\
      
  \hline
  
    \color{myred} $0-1$ \color{black} - $\lambda=8$ & \cellcolor{mypowderblue} $-0.08$ & \cellcolor{myblue} $\textbf{73.22}$ & \cellcolor{mypowderblue} $-0.24$ & \cellcolor{myblue} $\textbf{64.29}$ \\
      
  \hline

    \color{mygreen} $\frac{1}{2}-\frac{1}{2}$ \color{black} - $\lambda=8$ & \cellcolor{myblue} $\textbf{76.36}$ & \cellcolor{mypowderblue} $-0.44$ & \cellcolor{mypowderblue} $-0.14$ & \cellcolor{myblue} $\textbf{68.69}$ \\
    
  \hline
    
    \color{myred} $\frac{1}{2}-\frac{1}{2}$ \color{black} - $\lambda=8$ & \cellcolor{myblue} $\textbf{75.56}$ & \cellcolor{mypowderblue} $-0.37$ & \cellcolor{mypowderblue} $-0.32$ & \cellcolor{myblue} $\textbf{67.61}$ \\
      
  \hline
  
\end{tabular}
\caption{Pr\'ecision en pr\'esence de bruit}
\label{tableau:2}
\end{table}
\end{center}

\section{Exp\'eriences sur donn\'ees r\'eelles}

Les jeux de donn\'ees sont issus de l'UCI machine learning repository \cite{10}. Trois datasets sont utilis\'es : \textbf{Ionosphere} avec 351 donn\'ees de dimension 34, \textbf{Ecoli} avec 336 donn\'ees de dimension 8 et \textbf{Breast cancer} avec 699 donn\'ees de dimension 10.

\subsection{R\'esultats de classification}

Pour obtenir ces r\'esultats, 100 simulations ont \'et\'e effectu\'ees, et apr\`es 10 simulations successives, les \textit{train} et \textit{test set} sont recompos\'es (70\% train set et 30\% test set).

\begin{figure}[!h]
\centering
\subfigure[Ionosphere\label{fig:2a}]{\includegraphics[scale=0.21]{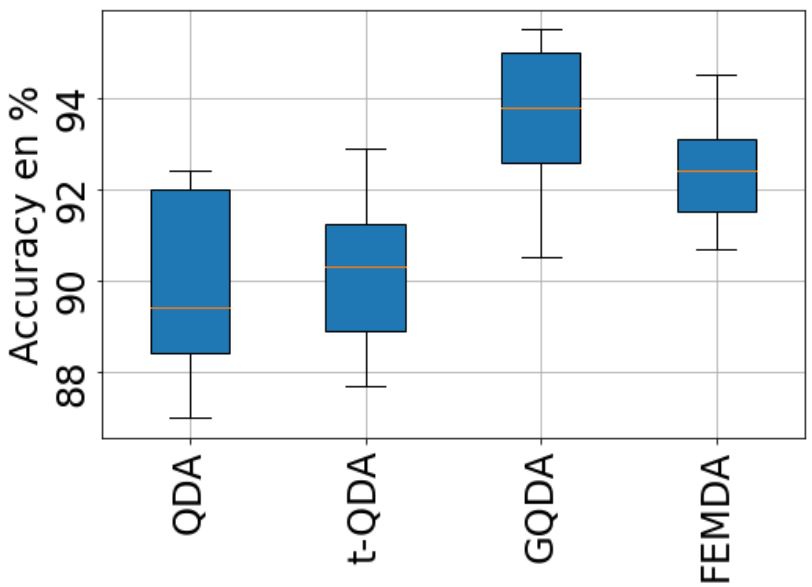}}
\subfigure[Ecoli\label{fig:2b}]{\includegraphics[scale=0.21]{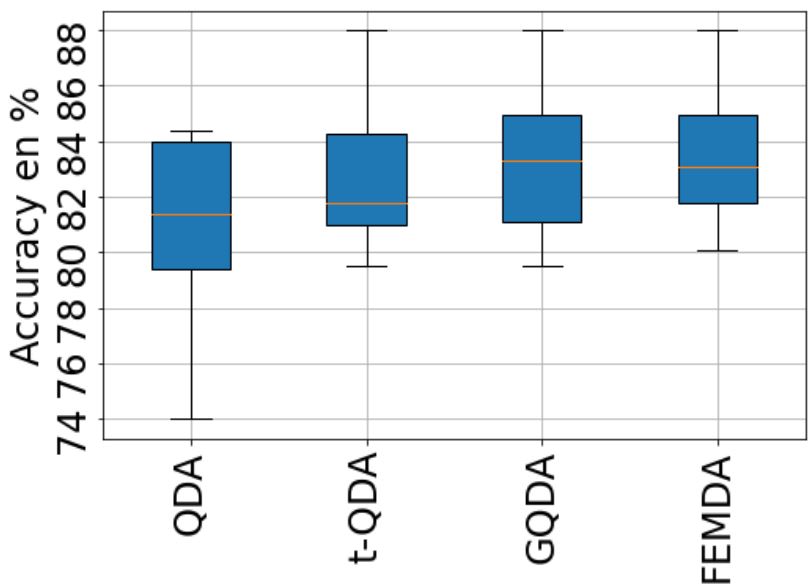}}
\subfigure[Breast cancer\label{fig:2c}]{\includegraphics[scale=0.21]{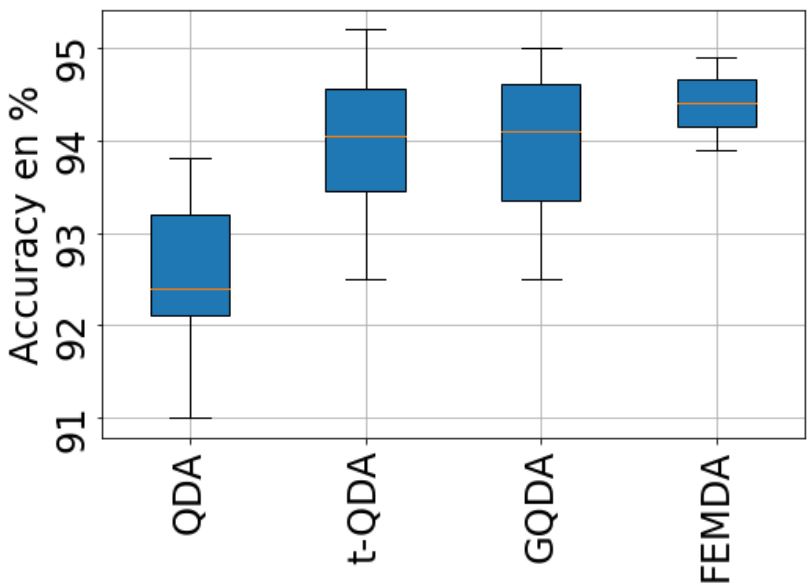}}
\caption{Pr\'ecision m\'ediane}
\label{fig:2} 
\end{figure}

On peut voir sur les figures \ref{fig:2a} et \ref{fig:2b} que GQDA surperforme d'au moins 1\% les autres m\'ethodes, suivi par FEMDA et ensuite par $t$-QDA pour le dataset Ionosphere. Les \'ecarts sont plus resserr\'es pour le dataset Ecoli. Sur la figure \ref{fig:2c}, on remarque que FEMDA devient la meilleure m\'ethode avec une pr\'ecision proche de 95\%, suivi de pr\`es par $t$-QDA puis GQDA. La variance dans les r\'esultats est plut\^ot faible. Pour conclure sur ces trois datasets, les performances de FEMDA sont l\'eg\`erement sup\'erieures \`a celles de $t$-QDA, et souvent inf\'erieures \`a celles de GQDA, qui sont cependant plus variables.

\subsection{R\'esultats apr\`es changements d'\'echelle}

On va maintenant bruiter les donn\'ees d'une mani\`ere similaire \`a ce qui a \'et\'e effectu\'e pour les donn\'ees simul\'ees. On choisit $\lambda=5$. On trace ensuite l'\'evolution de la pr\'ecision des trois m\'ethodes robustes selon le taux de contamination. 

\begin{figure}[!h]
\centering
\subfigure[Ionosphere\label{fig:3a}]{\includegraphics[scale=0.37]{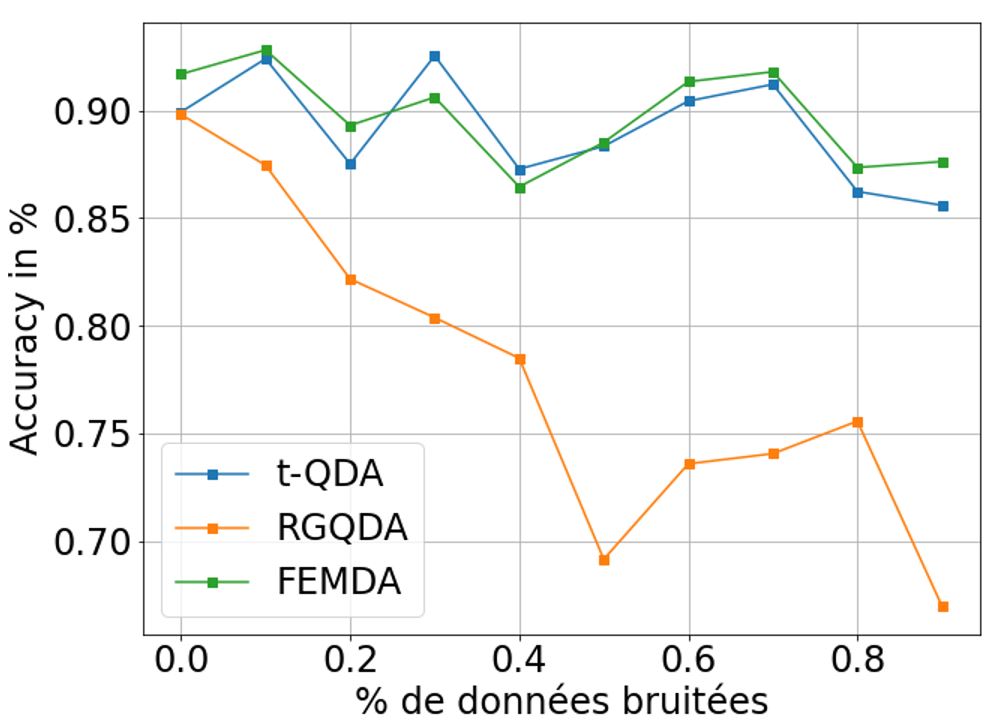}}
\subfigure[Ecoli\label{fig:3b}]{\includegraphics[scale=0.37]{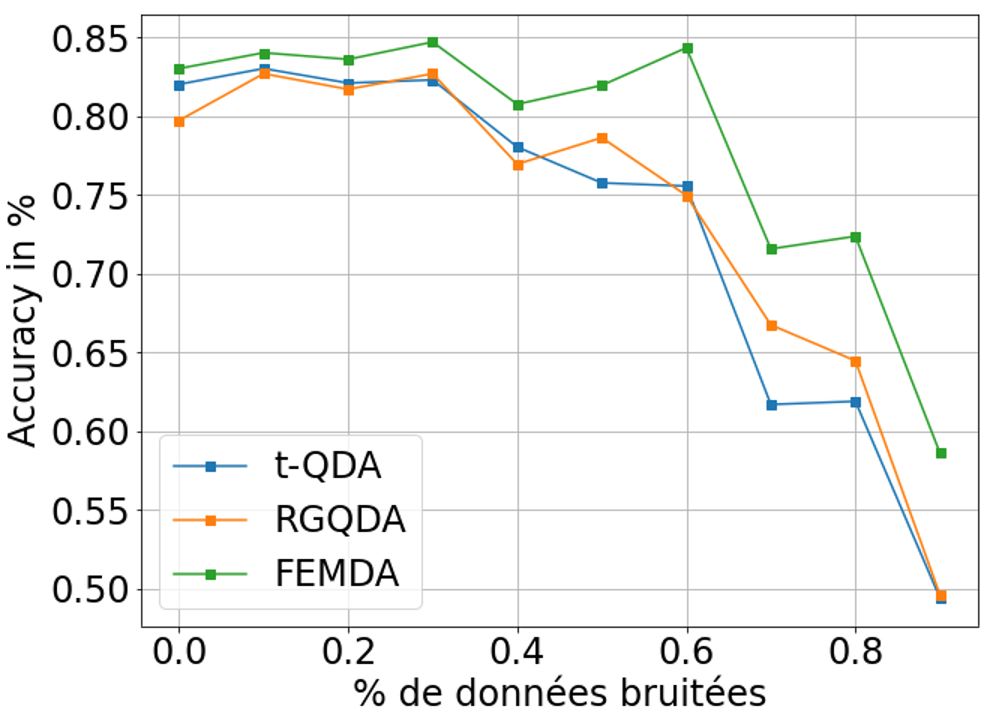}}
\subfigure[Breast cancer\label{fig:3c}]{\includegraphics[scale=0.37]{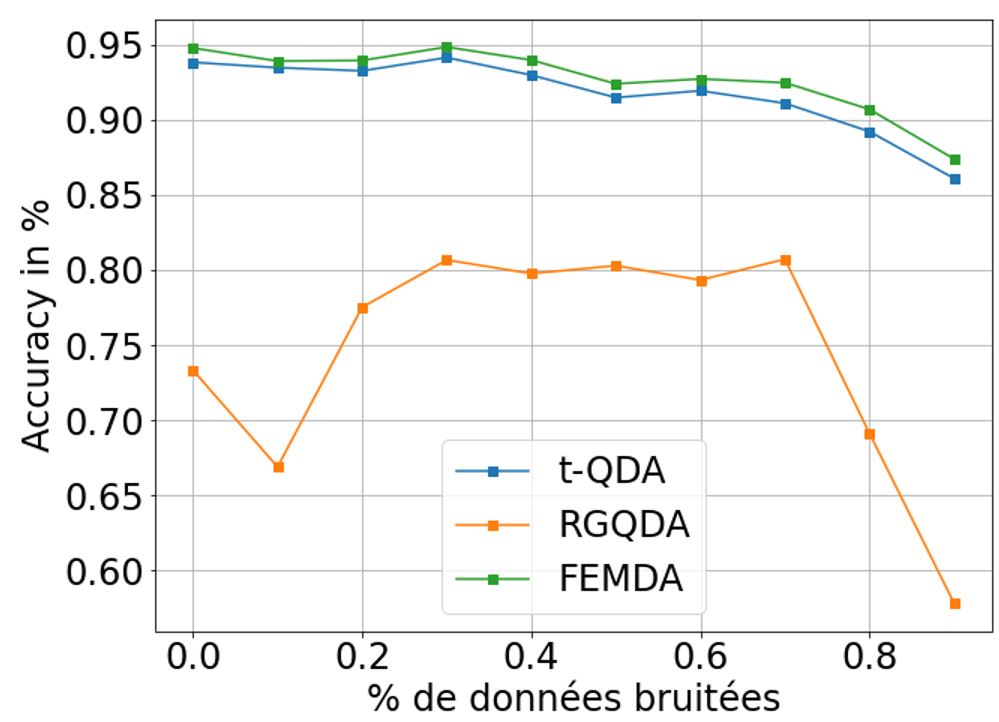}}
\caption{Donn\'ees bruit\'ees par changement d'\'echelle avec $\lambda=5$}
\label{fig:3} 
\end{figure}

On remarque sur la figure \ref{fig:3} que m\^eme avec des taux de bruit tr\`es \'elev\'es, $t$-QDA et FEMDA conservent de tr\`es bons r\'esultats pour Spambase et Ionosphere. En revanche, les performances de RGQDA chutent beaucoup plus rapidement lorsque le taux de contamination augmente.
Pour le dataset Ecoli, le comportement est beaucoup plus uniforme, les trois m\'ethodes voient leurs performances baisser, surtout lorsqu'on d\'epasse un taux de 40\% de contamination. FEMDA affiche malgr\'e tout une r\'esilience l\'eg\`erement sup\'erieure pour les hauts taux de contamination, mais les performances restent tr\`es proches de celles de $t$-QDA. 
La robustesse de FEMDA aux changements d'\'echelle dans les donn\'ees d'entra\^inement peut \^etre expliqu\'ee par l'expression des estimateurs, qui sont intrins\`equement insensibles aux changements d'\'echelle.
Enfin, la diff\'erence de sensibilit\'e \`a l'augmentation de la contamination pour Ecoli peut s'expliquer par la faible dimension des donn\'ees par rapport aux autres datasets. En effet, en grande dimension, la direction de la matrice de covariance est beaucoup plus discriminante pour s\'eparer les donn\'ees.

\section{Conclusion}
\label{sec:5}

Dans ce papier, nous avons pr\'esent\'e une nouvelle m\'ethode d'analyse discriminante robuste aux changements d'\'echelle dans les donn\'ees d'entra\^inement. Elle surpasse toutes les m\'ethodes de l'\'etat de l'art en pr\'esence de donn\'ees contamin\'ees, et se comporte de mani\`ere similaire \`a $t$-QDA sans bruit, tout en \'etant plus rapide. FEMDA est donc une m\'ethode rapide et tr\`es r\'esiliente face aux donn\'ees bruit\'ees. Dans ce nouveau paradigme, les clusters ne partagent plus la m\^eme matrice de covariance, mais seulement la m\^eme matrice de dispersion. Permettre \`a chaque point d'avoir son propre facteur d'\'echelle entra\^ine un gain de flexibilit\'e qui permet de traiter des jeux de donn\'ees contamin\'ees et non n\'ecessairement identiquement distribu\'ees. On peut donc consid\'erer que FEMDA est une am\'elioration de $t$-QDA : performances similaires sans contamination, mais plus robuste et plus rapide.

\end{document}